\documentclass[conference]{IEEEtran}
\IEEEoverridecommandlockouts
\usepackage{setspace}
\usepackage{cite}
\usepackage{flushend}
\usepackage{amsmath,amssymb,amsfonts}
\usepackage{algorithmic}
\usepackage{hyperref}
\usepackage{graphicx}
\usepackage{textcomp}
\usepackage{makecell}
\usepackage{xcolor}

\usepackage{colortbl}
\usepackage{natbib} 
\usepackage{url}

\usepackage{array} % Required for better tabular spacing control

\usepackage[table,xcdraw]{xcolor}
\usepackage{booktabs} % For formal tables
\usepackage{graphicx}  % for adjusting size of table
\usepackage{adjustbox} 
\usepackage{multirow}
\usepackage{caption}   % For adding captions
\usepackage{comment}

\usepackage{subcaption}
\usepackage{xcolor}
\def\BibTeX{{\rm B\kern-.05em{\sc i\kern-.025em b}\kern-.08em
    T\kern-.1667em\lower.7ex\hbox{E}\kern-.125emX}}
\usepackage{etoolbox}

\newbool{anonymous}
\setbool{anonymous}{true} % Mude para 'false' para ter autores
\newcommand{\citeAnonymous}[2]{%
    \ifbool{anonymous}{%
        \cite{#1}
    }{%
        \cite{#2}
    }%
}

\newcommand{\renderAuthors}{%
	{%
        \author{\IEEEauthorblockN{E. Riveros}
		\footnotesize \textit{Institute of Computing}\\
		\IEEEauthorblockA{\footnotesize \textit{State University of Campinas}\\
		{ \textit Unicamp} \\
		Campinas, Brazil\\
		e265685@g.unicamp.br}
		\and
		\IEEEauthorblockN{D. Vega-Oliveros}
		\footnotesize \textit{Institute of Science and Technology}\\
		\IEEEauthorblockA{\footnotesize \textit{Federal University of São Paulo}\\
		{ \textit UNIFESP} \\
		São José dos Campos, Brazil\\
		didier.vega@unifesp.br}
		\and 
		\IEEEauthorblockN{A. Soriano-Vargas}
		\footnotesize \textit{Dept. Académico de Ciencia de Computación y Datos}\\
		\IEEEauthorblockA{\footnotesize \textit{Universidad de Ingeniería y Tecnología} \\
		{ \textit UTEC} \\
		Lima, Perú \\
		asoriano@utec.edu.pe}
		\and
		\IEEEauthorblockN{A. Rocha}
		\footnotesize \textit{Institute of Computing}\\
		\IEEEauthorblockA{\footnotesize \textit{State University of Campinas} \\
		{ \textit Unicamp} \\
		Campinas, Brazil \\
		arrocha@unicamp.br}
		}
    }%
}

\begin{document}

%\title{Classification of Alimentary Activities Using CABiGRU: A BiGRU-Convolutional Model with Attention Mechanisms for Wearable Sensor Data}

%\title{CABiGRU: A BiGRU-Convolutional Model with Attention Mechanisms for Classifying Alimentary Human Activities}

\title{Diffusion-Based Synthetic Data Pretraining for Enhancing Activity Recognition}

\renderAuthors

% \author{\IEEEauthorblockN{1\textsuperscript{st} Omitted for double-blind Review}
% \IEEEauthorblockA{\textit{dept. name omitted for double-blind review} \\
% \textit{name of organization omitted for double-blind review}\\
% City, Country \\
% email address or ORCID}
% }

\IEEEaftertitletext{\vspace{-1.0\baselineskip}}
\maketitle

\begin{abstract}

Human activity recognition (HAR) is increasingly important for healthcare, well-being, and daily monitoring applications, for which detecting alimentary activities such as eating and drinking can provide actionable insight into dietary habits and chronic disease management. HAR systems, however, often underperform on subtle and underrepresented classes, limiting their utility in real-world dietary monitoring. This work builds upon CABiGRU, a convolutional architecture with Bidirectional GRU layers, multi-head attention, and residual connections, designed to capture discriminative temporal patterns from smartwatch accelerometer, gyroscope, and magnetometer data. To improve CaBiGRU's generalization and reduce underfitting in the minority class, we leverage synthetic sensor data windows using a diffusion model and adopt a two-stage training strategy: pre-training CABiGRU on synthetic data, followed by fine-tuning on the real-world data. On the DEO (drinking/eating/other) dataset, the proposed pipeline achieves a balanced accuracy of 90.6\%, improving over a strong supervised baseline and showing the benefits of diffusion-based synthetic pre-training for recognizing alimentary activities and representing a step forward dealing with unbalanced classes. These results suggest that combining diffusion-generated data with targeted fine-tuning enhances robust recognition of dietary behaviors, supporting more reliable deployment in healthcare and nutrition-monitoring settings.
\end{abstract}
\begin{comment}
Human activity recognition (HAR) has become paramount in healthcare, well-being, and daily activity monitoring applications. Identifying activities such as eating and drinking can provide valuable insights into dietary habits and the management of chronic diseases. However, existing HAR models often disregard minority classes, particularly in recognizing subtle and underrepresented alimentary activities, which are often overlooked and receive less emphasis in studies. In this context, we propose a deep-learning architecture for recognizing human alimentary activities based on smartwatch sensor data. It integrates convolutional layers for feature extraction, Bidirectional Gated Recurrent Unit layers to capture long-term dependencies in both directions, and multi-head attention mechanisms to emphasize the most relevant features. We also include residual connections to ensure feature retention, improving classification performance. Experimental results show that our method obtained the best classification output, with a balanced accuracy of 86.55\%, outperforming prior-art models. Our findings indicate that our approach enriches the recognition of complex alimentary activities, making it highly suitable for applications in healthcare and dietary monitoring.
\end{comment}

\section{Introduction}
Human Activity Recognition (HAR) is important for healthcare and daily monitoring, and wearable devices provide rich motion data for this purpose. Among HAR tasks, recognizing alimentary activities such as eating and drinking is especially relevant for dietary assessment and hydration monitoring. Unlike coarse-grained activities, these behaviors involve subtle wrist movements and short temporal patterns, making them harder to distinguish from smartwatch signals.

A key challenge in alimentary activity recognition is the combination of class imbalance and temporal subtlety. Eating and drinking are less frequent than other activities in real-world datasets, and their recognition depends on fine-grained wrist-motion dynamics and cross-sensor interactions. Prior work~\cite{riveros2025} introduced CABiGRU, achieving 86.6\% balanced accuracy, but minority-class scarcity still limits generalization.

To address this, augmentation strategies such as SMOTE, autoencoders, and GANs have been explored, but these methods do not always preserve the temporal continuity and multivariate dependencies required for subtle time-series discrimination. Diffusion models offer a more principled alternative, as they better model realistic multivariate temporal structure. In particular, PaD-TS~\cite{li2025} preserves population-level properties — including cross-correlation distributions — reducing distribution shift between real and synthetic data.

Based on this idea,  we propose a diffusion-based pretraining pipeline in which synthetic multivariate smartwatch data samples pre-train the recognition model before fine-tuning on real data. We further adapt PaD-TS by replacing attention in the encoder--decoder with Bidirectional GRU blocks, better suited to short inertial sequences. This recurrent redesign reduces overfitting and achieves 90.6\% balanced accuracy on the DEO dataset, outperforming both the 86.6\% scratch-trained baseline and adversarial synthetic pretraining.

Our contributions are: (1) introducing a diffusion-based pretraining strategy for fine-grained HAR, (2) developing architectures that improve temporal modeling by replacing attention with bidirectional GRUs, and (3) providing empirical validation on real-world wearable data, demonstrating improved accuracy for alimentary activity recognition.

We develop and test diffusion-model variants that use bidirectional GRU blocks instead of attention for denoising. This change better aligns with the short inertial sequences in our data. It improves temporal modeling and reduces overfitting when data is limited.

\textbf{Empirical evidence on real-world wearable data:}
We conduct a comprehensive evaluation using a smartwatch-based dataset with 90 participanst and a subject-wise protocol. Our approach reaches 90.6\% balanced accuracy. It outperforms both the strong baseline trained from scratch (86.6\%) and adversarial synthetic pretraining. This demonstrates the value of diffusion-based pretraining for fine-grained temporal classification.

\section{Related work}
\label{sec:related_work}

Wearable inertial sensors are widely used for HAR, with approaches ranging from classical models (SVMs, Random Forests) to deep learning on raw sensor streams~\cite{majidzadeh2020indirect,Nair2019,casale2011human,barut2020,singh2020,cheng2023}. Convolutional, recurrent, and self-supervised architectures have further improved recognition of spatial and temporal patterns~\cite{zheng2014,pamap2,wisdm,san2018robust,mahmud2020}. 
However, most work targets common daily activities; alimentary actions such as eating and drinking remain harder due to subtle wrist-motion dynamics and fine-grained temporal structure, as highlighted by systems such as AutoHydrate~\cite{mengistu2016} and multi-sensor drinking detection approaches~\cite{li2024}.

Synthetic time-series generation has emerged as a key strategy for data scarcity and class imbalance. GAN-based models such as PCF-GAN handle sequential data effectively~\cite{lou2023}, while diffusion-based models such as PaD-TS preserve population-level properties to reduce distributional shift~\cite{li2025}. However, these methods are typically evaluated for sample realism rather than downstream recognition utility.

In summary, prior work leaves three main gaps. First, most HAR models focus on common activities and do not explicitly address fine-grained eating--drinking distinctions. Second, alimentary-monitoring approaches often rely on handcrafted features, constrained protocols, or specialized sensing setups, which limits their applicability to smartwatch-based recognition. Third, although synthetic time-series generation has advanced substantially, it is still most often evaluated in terms of sample realism, distributional fidelity, or reconstruction quality, rather than its utility as a pretraining signal for downstream recognition. Our work addresses these gaps by investigating diffusion-based synthetic pretraining for fine-grained alimentary HAR. Specifically, we adapt the generator with Bidirectional GRU blocks for short multivariate smartwatch windows and evaluate the resulting pipeline under a subject-wise DEO protocol, comparing it against training from scratch and GAN-based synthetic pretraining.

\begin{figure*}[t]
\centering
\includegraphics[width=0.8\textwidth]{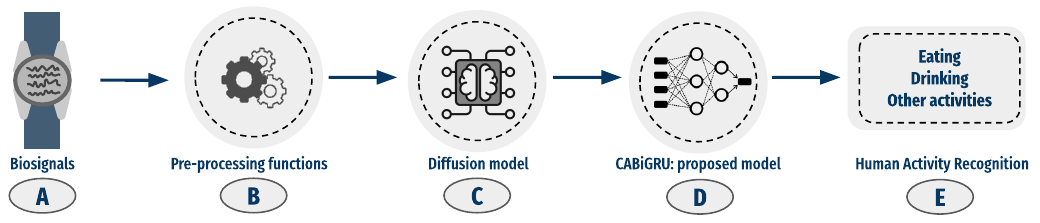}
\caption{Overview of the proposed pipeline for alimentary activity recognition: (A) acquisition of accelerometer, gyroscope, and magnetometer biosignals from a smartwatch; (B) preprocessing into 5-second windows with 50\% overlap; (C) diffusion model trained to generate synthetic multivariate windows; (D) CABiGRU trained with a two-stage strategy (pre-training on synthetic data and fine-tuning on real data); and (E) final classification into \textit{eating}, \textit{drinking}, or \textit{other} activities.}
\label{fig:overview}
\vspace*{-0.5cm}
\end{figure*}

\section{Proposed methodology} 
\label{sec:proposed_methodology}

In light of these recent advances, we aim to improve the recognition of alimentary activity under class imbalance regimes by increasing training diversity without altering the true test distribution. We propose a synthetic-data augmentation and two-stage training pipeline in which a generative model produces additional multivariate smartwatch windows; CABiGRU is first pre-trained on these synthetic samples and then fine-tuned on the original DEO data. This strategy increases exposure to plausible motion patterns while allowing the final model to adapt to real sensor characteristics and subject variability.

We segment accelerometer, gyroscope, and magnetometer signals into 5-second windows to provides enough temporal context to capture complete short wrist-motion patterns, with 50\% overlap to increase the training coverage and reduces the chance that brief eating or drinking gestures are split across segment boundaries. A diffusion model is trained on these multivariate windows to generate additional samples, preserving per-channel value distributions and cross-sensor dependencies to help reduce distribution shift.

Standard techniques such as SMOTE or autoencoders may not preserve the sequential nuances required for time-series signals, motivating the exploration of deep generative approaches for augmentation.

Following this augmentation, training proceeds in two steps: (1) pre-train CABiGRU on generated windows and (2) fine-tune on real DEO windows for task-specific adaptation. These steps improve the balanced accuracy and reduce overfitting compared to training from scratch. Our pipeline classifies smartwatch inertial biosignals into \textit{eating}, \textit{drinking}, and \textit{other} by segmenting and modeling raw streams using deep temporal architectures (see Figure~\ref{fig:overview}).

\subsection{Data Collection}
The Daily Living Activities Dataset used in this study includes 90 participants performing 13 common activities while wearing a Samsung GSW4 smartwatch on the wrist.The device records accelerometer, gyroscope, and magnetometer signals during 5-minute sessions. The cohort is predominantly composed of right-handed adults aged 20--30, with both male and female participants, as well as a smaller subset aged 50+. 

From this collection, we define the DEO dataset with three classes: drinking, eating, and other, where the last class groups all non-alimentary activities. This relabeling focuses the task on distinguishing fine-grained alimentary wrist-motion patterns while simplifying the recognition problem. For \textit{drinking} and \textit{eating}, data is taken from the participant's dominant hand, whereas for \textit{other}, data comes from the opposite hand.

Signals are sampled at 100 Hz and segmented into overlapping 5-second windows with 50\% overlap. Each window contains 500 time steps and 9 features per step, corresponding to the three axes of the accelerometer, gyroscope, and magnetometer.

\subsection{Diffusion-Based Synthetic Pretraining} 
After windowing the multivariate smartwatch signals, we train a diffusion model to generate additional accelerometer, gyroscope, and magnetometer samples for data augmentation. The model learns the joint distribution of the three sensors, preserving channel-wise statistics and cross-sensor dependencies, which helps reduce distributional shift between real and synthetic data.

Our approach is based on the Population-aware Diffusion for Time Series (PaD-TS) framework~\cite{li2025}. We modify PaD-TS by replacing its original attention-based encoder-decoder with Bidirectional GRU blocks, which are better suited for short multivariate inertial windows. This redesign maintains the population-aware objective while better capturing temporal dependencies and reducing overfitting.

The generated samples are used only during training and only for the minority classes, eating and drinking. This increases the diversity of realistic motion patterns in underrepresented classes and supports the two-stage training procedure described next. Unlike rebalancing methods such as SMOTE or autoencoders, diffusion-based augmentation generates full multivariate time-series windows, making it better suited to preserving the temporal dependencies needed to distinguish between eating and drinking.

\subsection{CABiGRU Backbone} 
    We adopt CABiGRU as the downstream recognition backbone in this work because it was previously proposed and validated for smartwatch-based alimentary activity recognition, where it achieved strong performance on the DEO dataset. CABiGRU combines convolutional layers for feature extraction, bidirectional GRU layers for temporal modeling, multi-head attention, and residual connections, making it a suitable baseline for subtle wrist-motion classification. 
   It is worth noting that this choice does not restrict the generality of the proposed method. The main contribution of this work is the diffusion-based synthetic pretraining strategy, which is conceptually independent of the specific recognition backbone and could be integrated with other sequence models for multivariate inertial HAR. We use CABiGRU here because it is a strong previously validated architecture for this task, allowing us to assess the effect of synthetic pretraining under a competitive and relevant downstream model.

\subsection{Two-Stage Training (Pre-train + Fine-tune)}
In our two-stage training process, the first phase trains only the convolutional and GRU layers of CABiGRU on generated sensor data, acting as self-supervised pre-training. Since the generated data is class-specific, the model captures general temporal and cross-sensor patterns without overfitting to the small real dataset.

In the fine-tuning stage, the classification head is introduced and the full model is optimized solely on real data using a lower learning rate (1e-4), enabling gradual adaptation of the pretrained representations while reducing overfitting risk.

To evaluate the impact of architectural design, we test four classifier head configurations — balanced (BatchNorm + Dense(256, 128) + Dropout), light (single Dense + Dropout), deep (BatchNorm + Dense(512, 256) + higher Dropout), and classic (Dense(100)) — combined with four activation functions: GELU, SiLU, Tanh, and ReLU. Regardless of mode, the final layer is always a softmax Dense layer for class probability output.

Among all combinations, the best result was achieved with GELU + \texttt{deep} head (balanced accuracy of 0.9055), followed by Tanh + \texttt{classic} (0.8986) and Tanh + \texttt{balanced} (0.8964). Overall, the top configurations consistently exceeded 0.89 balanced accuracy, demonstrating that the \texttt{deep} and \texttt{balanced} heads paired with smooth activation functions (GELU, Tanh) yield the most robust representations.

This two-stage procedure improves balanced accuracy and reduces overfitting compared to training CABiGRU from scratch on the original dataset. In the next section, we describe the experimental setup and evaluation protocol used to validate this pipeline, connecting our methodological innovations to their empirical validation.

% \begin{figure}[t]
% \centering
% \includegraphics[[width=0.8\textwidth]{img/DF-BiGRU}
% \caption{A diffusion model inspired by~\cite{li2025} is trained on raw multivariate time-series windows using Bidirectional GRU layers.}
% \vspace*{-0.8cm}
% \label{fig:difusion_model}
% \end{figure}

\begin{figure}[htbp]
    \centering
    \includegraphics[width=1.2\linewidth]{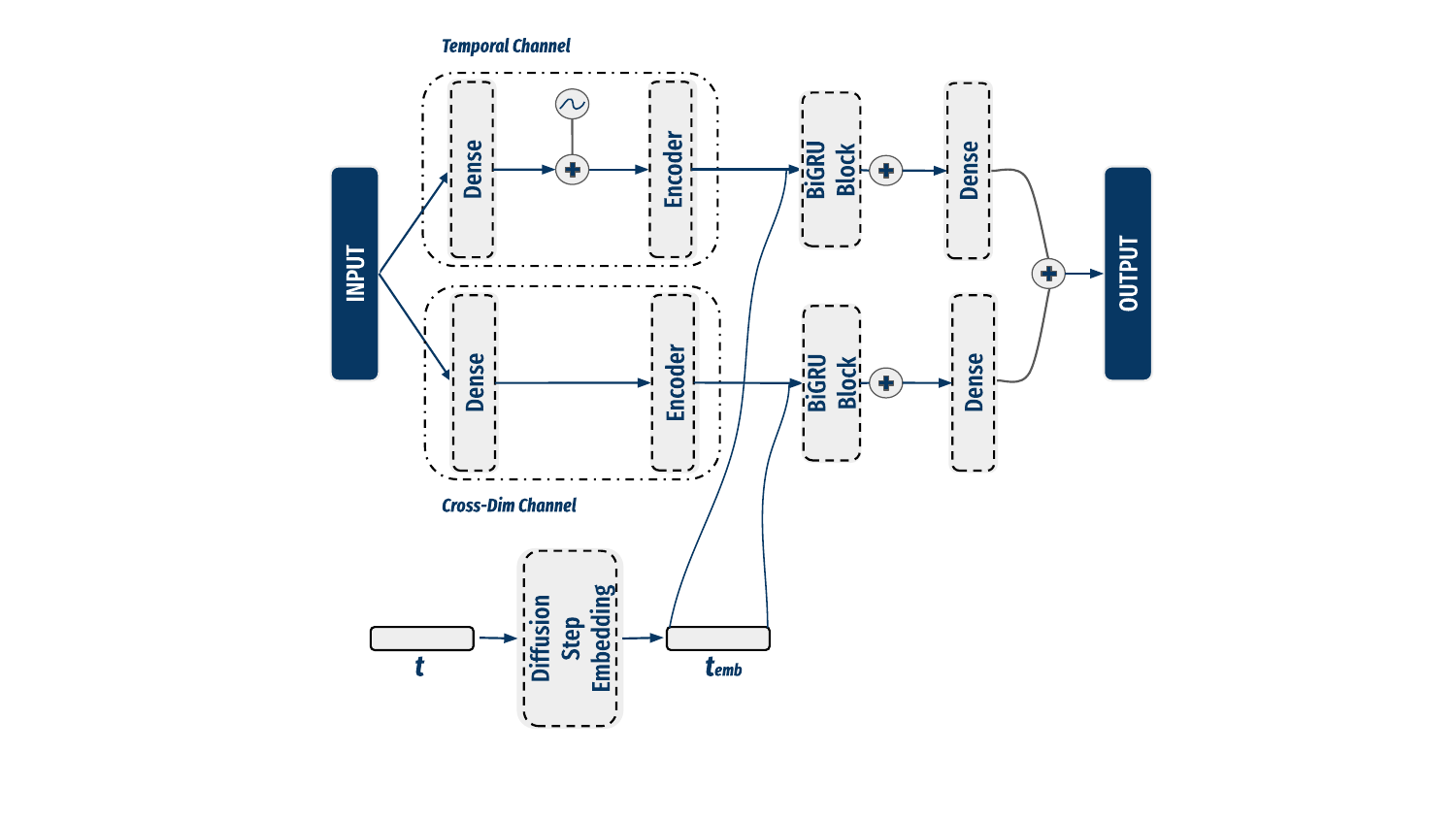}
    \caption{A diffusion model inspired by~\cite{li2025} is trained on raw multivariate time-series windows using Bidirectional GRU layers.}
    \label{fig:diffusion_model}
\end{figure}

% \begin{figure*}[tb!]
% \centering
% \includegraphics[width=0.6\textwidth]{img/proposed_model.png}
% \caption{Proposed Model Architecture: Two stacked convolutional blocks for feature extraction, followed by Bidirectional Gated Recurrent Unit (BiGRU) layers that capture long-term dependencies in human activity data. Residual connections and attention enhance stability and relevance during training, yielding a fixed-size representation, while regularization prevents overfitting.}
% \label{fig:proposed_model}
% \end{figure*}

% \begin{figure}[tb!]
% \centering
% \includegraphics[width=0.4\textwidth, height=0.3\textwidth, angle=-90]{img/proposed_model.png}
% \caption{Proposed Model Architecture: Two stacked convolutional blocks for feature extraction, followed by Bidirectional Gated Recurrent Unit (BiGRU) layers that capture long-term dependencies in human activity data. Residual connections and attention enhance stability and relevance during training, yielding a fixed-size representation, while regularization prevents overfitting.}
% \label{fig:proposed_model}
% \end{figure}

\section{Experimental Results}
\label{sec:experiments_results}

This section describes the experimental protocol and reports the performance of our diffusion-based synthetic pretraining pipeline. We outline the evaluation setup and metrics, then present quantitative results and analyses on the DEO multi-class dataset.

% \\\\\\\\\\\\\\\\\\\\\\\\\\\\\\\\\\\\\\\\\\\\\\\\\\\\\\\\\\\\\\\\\%
\subsection{Experimental Setup and Evaluation Protocol}
\label{subsec:setup_evaluation}
% Section~IV-A

We evaluate all models using a strict subject-wise protocol. The dataset is partitioned by subjects, allocating 60\% for training, 10\% for validation, and 30\% for testing. This split balances learning capacity and evaluation robustness: a relatively large, subject-disjoint test set enables reliable out-of-sample performance estimation, while the training portion remains sufficiently large to support both the generative model and the downstream classifier.

Synthetic samples are used solely for self-supervised pretraining, where CABiGRU learns general representations from augmented windows. The model is then fine-tuned for classification on the original train/validation/test splits, avoiding information leakage since synthetic data is generated from eating and drinking classes.

Given the class imbalance and subtle nature of alimentary activities, we report balanced accuracy (mean of per-class sensitivity and specificity) as the primary metric, weighting minority and majority classes equally. We additionally report per-class sensitivity, specificity, confusion matrix, ROC curves (AUC per class), and learning curves.

% \\\\\\\\\\\\\\\\\\\\\\\\\\\\\\\\\\\\\\\\\\\\\\\\\\\\\\\\\\\\\\\\\%
\subsection{Diffusion-Based Synthetic Pretraining}
We first report the results of our main approach: CABiGRU trained using the proposed two-stage strategy. This strategy consists of pretraining on diffusion-generated synthetic windows, followed by fine-tuning on the original real training data. Under this protocol, the model achieves a test balanced accuracy of 0.9055, reflecting consistent performance across classes despite pronounced imbalance. Per-class results confirm strong recognition across all activities: sensitivity values are 0.8849 (Drink), 0.9014 (Eat), and 0.9304 (Other), with specificities of 0.9486, 0.9779, and 0.9537, respectively. These results support the hypothesis that exposure to diverse yet structurally coherent synthetic sequences enhances the model's ability to capture fine-grained temporal patterns, particularly for minority alimentary classes.

To identify which classes are most frequently confused, we analyze the confusion matrix and ROC curves. The confusion matrix analysis shows that most misclassifications occur between Drink and Eat, which is expected given their overlapping wrist-motion dynamics, while the Other class remains highly separable. ROC curves corroborate this, with AUC values ranging from 0.92 to 0.94 across all three classes.

\begin{figure}[htbp]
\centering
    \begin{subfigure}[t]{0.5\linewidth}
        \includegraphics[width=\linewidth]{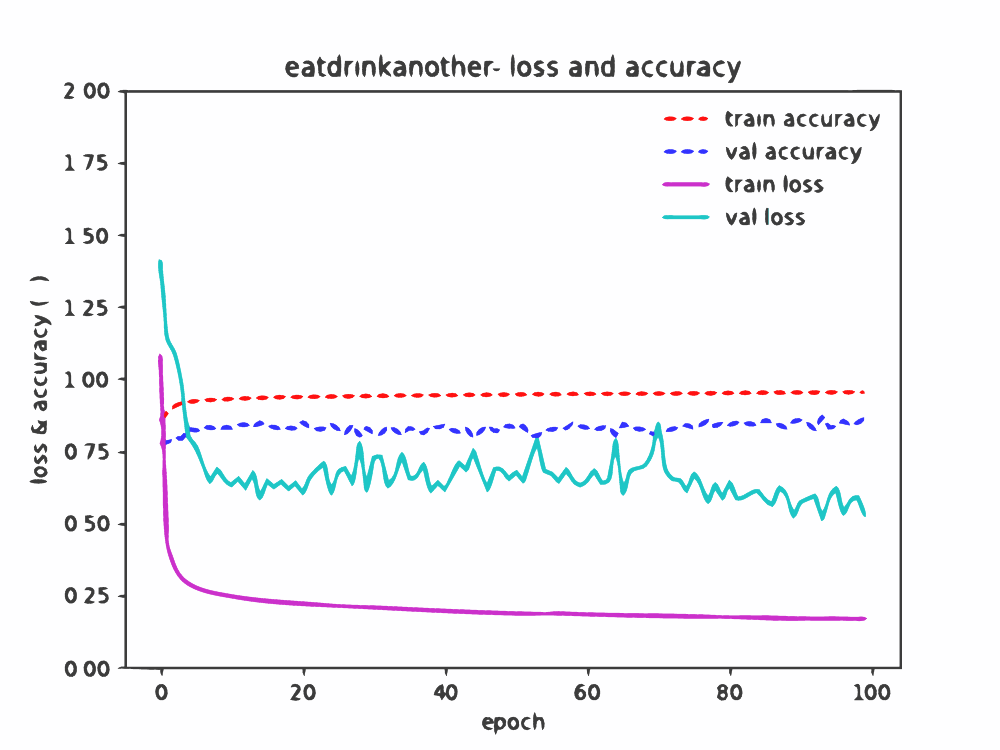}
        \caption{Pretraining without synthetic data.}
        \label{fig:loss_no_pretrain}
    \end{subfigure}\hfill
    \begin{subfigure}[t]{0.5\linewidth}
        \includegraphics[width=\linewidth]{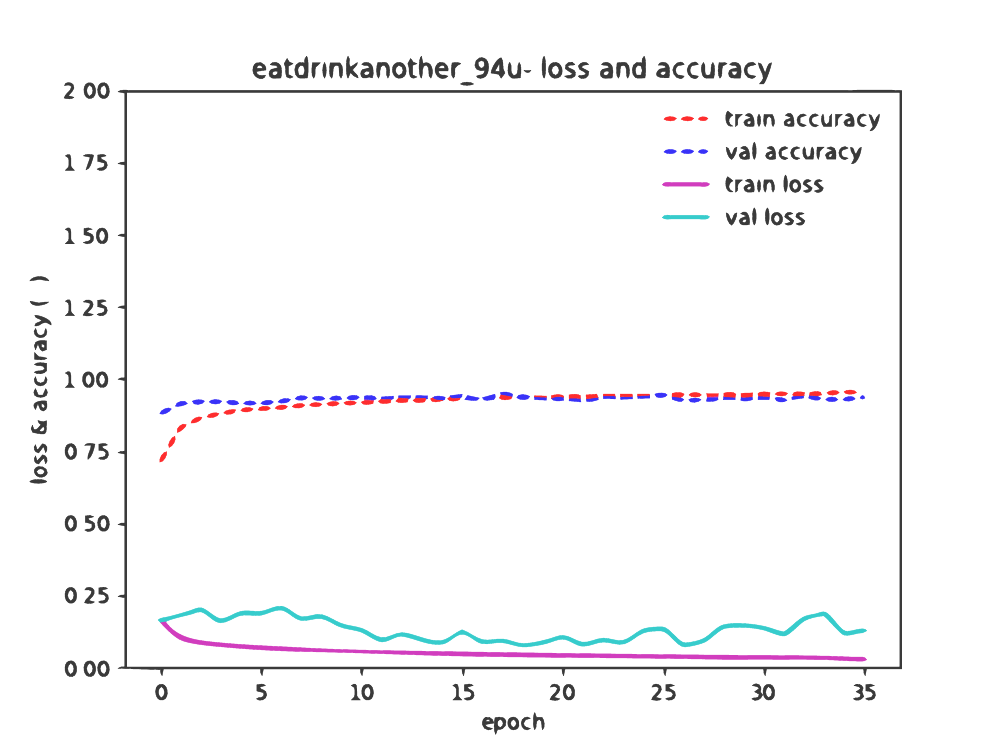}
        \caption{Pretraining with synthetic-data augmentation.}
        \label{fig:loss_pretrain}
    \end{subfigure}
    \caption{Training and validation accuracy and loss curves for CABiGRU under two training settings. The pretraining-based strategy yields more stable convergence, lower validation loss, and a smaller train-validation gap, indicating reduced overfitting and improved generalization.}
    \label{fig:accuracy_plots}
    \vspace*{-0.4cm}
\end{figure}

Learning curves further illustrate the benefit of pretraining (Fig.~\ref{fig:accuracy_plots}). Without it (Fig.~\ref{fig:loss_no_pretrain}), the model overfits quickly, showing an unstable validation loss and a larger train-validation gap. With synthetic pretraining (Fig.~\ref{fig:loss_pretrain}), training and validation curves converge more consistently and validation loss remains lower and stable, qualitatively supporting the quantitative gains reported in Table~\ref{tab:diffusion_vs_pcfgan}.

% \begin{figure}[htbp]
% \centering
%     \begin{subfigure}[t]{0.5\linewidth}
%         \includegraphics[width=\linewidth]{img/plot_loss_accuracy_0.85.png}
%         \caption{Pretraining without synthetic data.}
%         \label{fig:loss_no_pretrain}
%     \end{subfigure}\hfill
%     \begin{subfigure}[t]{0.5\linewidth}
%         \includegraphics[width=\linewidth]{img/plot_loss_accuracy_0.9.png}
%         \caption{Pretraining with synthetic-data augmentation.}
%         \label{fig:loss_pretrain}
%     \end{subfigure}
%     \caption{Training and validation accuracy and loss curves for CABiGRU under two training settings. The pretraining-based strategy yields more stable convergence, lower validation loss, and a smaller train-validation gap, indicating reduced overfitting and improved generalization.}
%     \label{fig:accuracy_plots}
%     \vspace*{-0.4cm}
% \end{figure}

\subsection{PCF-GAN Synthetic Pretraining }
As an adversarial baseline, we evaluate PCF-GAN for synthetic pretraining under the same subject-wise protocol used for diffusion. Additional synthetic windows are generated only for the minority classes (Drinking and Eating) and used only during the initial pre-training stage. CABiGRU is then fine-tuned on the original real DEO training split, while validation and test sets remain fully real. This baseline allows us to compare diffusion-based pretraining against an alternative generative strategy for sequential data under matched downstream conditions.
% To compare diffusion-based generation with an adversarial alternative, we evaluate a Generative Adversarial Network (PCF-GAN). PCF-GAN is a model for sequential data that uses the path characteristic function in the discriminator to better capture temporal dependencies. In this baseline, we generate additional synthetic windows for the minority alimentary classes (Drinking and Eating) and use them only in the initial pre-training stage. After pre-training, CABiGRU is fine-tuned on the original (real) DEO training split. The same subject-wise protocol is used. Validation and test sets remain fully real.

% We first assess the quality of generated samples using distributional similarity and embedding-based diagnostics. Using Sig-MMD, we get a mean of 0.00347 (std 0.0). This indicates a small discrepancy between real and synthetic sequences under this metric. We also analyze the overlap between real and synthetic samples in a low-dimensional representation. This analysis yields a mean t-SNE sigma of 0.7061.

\subsection{Comparative Discussion and Error Analysis}
We compare three strategies: CABiGRU trained from scratch (baseline), diffusion-based pretraining followed by fine-tuning, and PCF-GAN pretraining followed by fine-tuning. Diffusion achieves the best test balanced accuracy (0.9055), outperforming PCF-GAN (0.8802) and the from-scratch baseline, indicating that diffusion-generated windows provide more effective support for learning under class imbalance.

This gap reflects the nature of the generated sequences rather than data volume alone. In fine-grained HAR tasks such as eating versus drinking, class discrimination depends on subtle temporal ordering and coordinated multisensor dynamics. Diffusion's iterative denoising enforces local temporal consistency and cross-sensor dependencies at each step, whereas adversarial approaches match global distributions and may miss the precise temporal structure needed to distinguish subtle activities.

Among all tested combinations of activation functions (GELU, SiLU, Tanh, ReLU) and classifier head configurations (light, balanced, deep, classic), GELU with the \texttt{deep} head achieved the best balanced accuracy (0.9055) and was used in the main comparative experiments.

% Elian: This paragraph was removed because it is not central to the paper’s main claim.
% Table~\ref{table:best_balanced_accuracy_split} presents the top 10 results for balanced accuracy when combining activation functions and classifier head types. The table highlights the balanced accuracy for each configuration. Gelu activation with the 'deep' classifier head type achieves the highest balanced accuracy of 0.9055. This combination stands out for the smooth nonlinearity of gelu, which improves gradient flow and enables the model to learn more expressive features. The 'deep' classifier head, with its complex architecture including BatchNormalization, Dense(512), Dense(256), and stronger Dropout, supports better modeling of intricate patterns in the data. This enhances performance, especially for complex and subtle alimentary activities like eating and drinking. This result suggests that a deeper classifier head paired with gelu activation provides the best trade-off between capacity and generalization. This leads to superior performance in recognizing these activities.

Table~\ref{tab:diffusion_vs_pcfgan} summarizes downstream classification performance after fine-tuning. On the test set, the PCF-GAN pretraining baseline reaches a balanced accuracy of 0.8802, with per-class sensitivities of 0.8139 (Drink), 0.9216 (Eat), and 0.9051 (Other), and specificities of 0.9526, 0.9500, and 0.9333, respectively. While PCF-GAN improves minority-class detection over the from-scratch baseline, its gains remain below diffusion, suggesting that adversarial generation is less effective at preserving the temporal continuity and multivariate coherence required for alimentary HAR.

Diffusion-based pretraining also achieves a higher macro F1 (0.8243) than PCF-GAN (0.7764), reflecting more consistent improvement across classes and particularly for underrepresented alimentary categories. This confirms that diffusion-generated windows better enrich the minority-class training distribution rather than simply boosting majority-class performance.

Finally, diffusion also outperforms PCF-GAN in weighted F1 (0.9315 vs. 0.9089), indicating stronger overall performance across the full test set. Together, the macro and weighted F1 results suggest that diffusion-based pretraining achieves a better balance between minority-class sensitivity and overall classification robustness.

% \begin{table}[htbp]
% \centering
% \caption{Top 10 Best Balanced Accuracy Results with Split Hyperparameters}
% \renewcommand{\arraystretch}{1.0}
% \setlength{\tabcolsep}{10pt}
% \begin{tabular}{cccc}
% \hline
% \textbf{\#} & \textbf{Balanced} & \multicolumn{1}{c}{\textbf{Activation}} & \multicolumn{1}{c}{\textbf{Classifier}} \\
% &  \textbf{Accuracy} & \textbf{Function} & \textbf{Head Type} \\
% \hline
% 1  & \textbf{0.9055} & gelu & deep \\
% 2  & \textbf{0.8986} & tanh & classic \\
% 3  & \textbf{0.8964} & tanh & balanced \\
% 4  & \textbf{0.8952} & relu & deep \\
% 5  & \textbf{0.8945} & silu & light \\
% 6  & \textbf{0.8943} & silu & balanced \\
% 7  & \textbf{0.8935} & tanh & deep \\
% 8  & \textbf{0.8925} & relu & balanced \\
% 9  & \textbf{0.8924} & gelu & balanced \\
% 10 & \textbf{0.8907} & gelu & classic \\
% \hline
% \end{tabular}
% \label{table:best_balanced_accuracy_split}
% \end{table}

\begin{table}[t]
\centering
\footnotesize
\caption{Comparison of downstream task performance under different data generation strategies for the minority classes of DEO \textit{drink--eat} using a subject-wise split. The first column reports the baseline performance without any data augmentation.}
\label{tab:diffusion_vs_pcfgan}
\begin{tabular}{lccc}
\hline
\textbf{Metric} & \makecell{\textbf{No synthetic}\\\textbf{data}} & \makecell{\textbf{PCF-GAN}\\\textbf{pretrain}} & \makecell{\textbf{Diffusion}\\\textbf{pretrain}} \\
\hline
Balanced accuracy & 0.8655 & 0.8802 & 0.9055 \\
\hline
Sensitivity (Drink)  & 0.8269 & 0.8139 & 0.8849 \\
Sensitivity (Eat)    & 0.8701 & 0.9216 & 0.9014 \\
Sensitivity (Other)  & 0.8995 & 0.9051 & 0.9304 \\
\hline
Specificity (Drink)  & 0.9404 & 0.9526 & 0.9486 \\
Specificity (Eat)    & 0.9516 & 0.9500 & 0.9779 \\
Specificity (Other)  & 0.9491 & 0.9333 & 0.9537 \\
\hline
F1 (macro)           & 0.7573 & 0.7764 & 0.8243 \\
F1 (weighted)        & 0.9035 & 0.9089 & 0.9315 \\
mAP score            & 0.6427 & 0.6653 & 0.7272 \\
\hline
\end{tabular}
\vspace{-0.5cm}
\end{table}

\subsection{Effect of Synthetic Data Volume}

% To investigate whether increasing the volume of synthetic data further improves generalization, we conduct a scaling experiment on the diffusion-based pretraining pipeline. Starting from the baseline configuration in which $N$ synthetic windows are generated per minority class (matching the size of the original training split), we progressively double the synthetic set to $2N$, $4N$, and so on. At each scale, CABiGRU is pre-trained on the corresponding synthetic set and then fine-tuned on the original real DEO training split under the same subject-wise protocol described in Section~\ref{subsec:setup_evaluation}. We stop when the test's balanced accuracy no longer improves over the previous scale.

% Table III reports balanced accuracy as a function of synthetic data volume. The results show that increasing the synthetic data volume from N (baseline) to  2N leads to a decrease in balanced accuracy, from 0.9055 to 0.8338. This suggests that, in this setting, adding more diffusion-generated minority-class windows did not improve downstream performance and may have introduced redundant or less informative synthetic samples.

To investigate whether increasing synthetic data volume improves generalization, we conducted a scaling experiment starting from the baseline configuration of $N$ synthetic windows per minority class, then doubling it to $2N$. For each setting, CABiGRU was pre-trained on diffusion-generated windows and fine-tuned on the original real DEO training split under the same subject-wise protocol (Section~\ref{subsec:setup_evaluation}).

Increasing the synthetic pretraining volume from $N$ to $2N$ reduced test balanced accuracy from 0.9055 to 0.8338, indicating that the benefit of synthetic pretraining is not monotonic with data volume. Beyond a certain point, the generator likely produces redundant or synthetic-specific patterns rather than useful diversity, reducing the transferability of the pretrained representation to the real DEO distribution. This is consistent with validation-based model selection during fine-tuning, indicating that the observed degradation is more likely due to reduced transferability of the $2N$ pretrained representation than to unstable downstream optimization.

\section{Conclusions and Future Work}
\label{sec:conclusion_future_work}

In this study, we proposed a two-stage training pipeline in which CABiGRU is first pre-trained on diffusion-generated data samples and then fine-tuned on real smartwatch signals. Synthetic samples are generated only for the minority classes, eating and drinking receive greater representation during pretraining, which reduces the original class imbalance and helps the model learn richer temporal representations for subtle wrist-motion patterns. This strategy leads to improved generalization and robustness, especially for subtle and underrepresented activities such as eating and drinking.

The fine-tuning phase on real data let the model adapt to specific sensor signals, resulting in superior performance. Using a low learning rate ($1e-4$) and various activation functions and classifier head architectures enabled efficient convergence. This led to a significant boost in balanced accuracy, especially for subtle activities such as eating and drinking.

A scaling analysis further showed that synthetic augmentation must be used carefully. In fact, larger sets reduced balanced accuracy. When the synthetic pretraining volume was increased from 
$N$ to $2N$, balanced accuracy dropped from $0.9055$ to $0.8338$, indicating that synthetic data augmentation is important, but it should not be overused.

In our setting, excessive synthetic data may introduce redundant or less transferable patterns, which weakens the benefit of pretraining on real downstream recognition.
Our results show that diffusion models preserve temporal continuity and cross-sensor dependencies, especially when the target classes are subtle and underrepresented.

Future research will expand this approach to more diverse datasets, including those of older adults and individuals with specific health conditions. We also plan to add more sensor modalities, refine feature extraction, and develop advanced explainability methods to further improve performance and transparency in alimentary activity recognition.

%\section*{References}
\bibliographystyle{plain} % We choose the "plain" reference style
\small
\bibliography{references}
\end{document}